\documentclass[10pt,twocolumn,letterpaper]{article}
\usepackage{multirow}
\usepackage{wacv}
\usepackage{times}
\usepackage{epsfig}
\usepackage{graphicx}
\usepackage{amsmath}
\usepackage{amssymb}
\usepackage{graphicx}
\usepackage{tabularx}
\usepackage{subfig}
\usepackage{wrapfig,lipsum,booktabs}
\usepackage{algorithm}
\usepackage{algpseudocode}
\usepackage{algorithmicx}
\usepackage{url}
\usepackage{color}

\wacvfinalcopy 
\def\wacvPaperID{482} 

\ifwacvfinal\fi
\begin{document}

\title{Stability Based Filter Pruning for Accelerating Deep CNNs}

\author{Pravendra Singh\\
IIT Kanpur\\
{\tt\small psingh@iitk.ac.in}
\and
Vinay Sameer Raja Kadi \hspace{1cm} Nikhil Verma \\
Samsung R\&D Institute, Delhi\\
{\tt\small \{k.raja, nikhil.v\}@samsung.com}
\and
Vinay P. Namboodiri\\
IIT Kanpur\\
{\tt\small vinaypn@iitk.ac.in}
}

\maketitle
\ifwacvfinal\thispagestyle{empty}\fi

\begin{abstract}
  Convolutional neural networks (CNN) have achieved impressive performance on the wide variety of tasks (classification, detection, etc.) across multiple domains at the cost of high computational and memory requirements. Thus, leveraging CNNs for real-time applications necessitates model compression approaches that not only reduce the total number of parameters but reduce the overall computation as well. In this work, we present a stability-based approach for filter-level pruning of CNNs. We evaluate our proposed approach on different architectures (LeNet, VGG-16, ResNet, and Faster RCNN) and datasets and demonstrate its generalizability through extensive experiments. Moreover, our compressed models can be used at run-time without requiring any special libraries or hardware. Our model compression method reduces the number of FLOPS by an impressive factor of 6.03X and GPU memory footprint by more than 17X, significantly outperforming other state-of-the-art filter pruning methods.
\end{abstract}

\section{Introduction}
   In recent years, CNNs (convolutional neural networks) are being widely used in areas such as vision, NLP and other domains. While CNNs exhibit superior performance on a wide variety of tasks, their deployment calls for high-end devices due to their intensive computation (FLOPS) and memory requirements. This hinders their real-time usage on portable devices. While it may seem straightforward to address this problem by using smaller sized networks, redundancy of parameters seems necessary in aiding highly non-convex optimization during training to find effective solutions. Hence significant efforts are seen in recent days to address model compression. One line of research aims at devising efficient architectures \cite{huang2017condensenet,iandola2016squeezenet} to be trained from scratch on a given task. While they have shown promising results, their generalizability across the tasks is not fully studied. Another prominent line of work \cite{li2016pruning} has focused on model compression to make CNNs more efficient in terms of computations (FLOPS) and memory requirements (Run Time Memory usage and storage space of the model). These methods first train a large model for a given task and then prune the model until the desired compression is achieved.
   
   Model compression techniques can be broadly divided into the following categories. The first category  \cite{chen2015compressing, han2015deep} aims at introducing sparsity in the parameters of the model. While these approaches achieved good compression rate in model parameters, computations (FLOPS) and Total Runtime Memory (TRM) aren't improved. Such methods also require sparse libraries support to achieve the desired compression as mentioned in \cite{han2015deep}
   
The second category of methods \cite{han2015deep,louizos2017bayesian,binarycompression} based on model compression using quantization. Often specialized hardware is required to achieve the required acceleration. These model compression techniques are specially designed for IoT devices. 

The third category of methods \cite{abbasi2017structural,denton2014exploiting,he2018soft,li2016pruning,luo2017thinet,zhang2015efficient} based on the filter level pruning in the model. These approaches prune an entire filter based on some criteria/metrics and hence provide a structured pruning in the model. As for pruning the whole convolutional filter from the model reduces the depth of the feature maps for subsequent layers, these approaches give high compression rate regarding computations (FLOPS) and Total Runtime Memory (TRM). Moreover, sparsity and quantization based methods can be applied in addition to these approaches to achieve better compression rates.

As described above, filter level pruning approaches use a metric to identify the filter importance, and many heuristics have been used to identify the filter importance. \cite{abbasi2017structural} used the brute force approach to prune the filters from the model. They remove each filter sequentially and rank the importance of the filter based on their corresponding drop in the accuracy which seems to be impractical for large size networks on large-scale data-sets. Some of the works \cite{li2016pruning,luo2017entropy} use handcrafted metrics to calculate the filter importance. In the work of \cite{li2016pruning} they use ${l_1}$ norm of a filter to identify the filter importance. Another class of works \cite{luo2017thinet,molchanov2016pruning,ye2018rethinking} use data-driven metrics to identify the filter importance. \cite{molchanov2016pruning} use the Taylor expansion to calculate the filter importance, which is motivated by optimal brain damage \cite{hassibi1993second,lecun1990optimal}. 

In this work, we propose a new method for filter level pruning based on the sensitivity of filters to auxiliary loss function. While most of the pruning methods use sparsity in some form, our approach is orthogonal to them by choice of auxiliary loss function (by driving filter values away from 0) in the model. We evaluate our approach on a variety of tasks and show an impressive reduction in FLOPS across different architectures. We further demonstrate the generalizability of our approach by achieving competitive accuracy using a small model pruned for a different task. To further decrease the FLOPS and memory consumption, our method can be augmented with other pruning methods such as quantized weights, specialized network architectures devised for embedded devices, connection pruning,  etc.

\section{Related Work}
The works on model compression can be divided into the following categories. 
\subsection{Connection Pruning}
In the connection pruning, they introduce sparsity in the model by removing unimportant connections (parameters). There are many heuristics proposed to identify the unimportant parameters. Earliest works include Optimal Brain Damage  \cite{lecun1990optimal} and Optimal Brain Surgeon \cite{hassibi1993second} where they used Taylor expansion to identify the parameters significance. Later \cite{han2015deep} proposed an iterative method where absolute values of weights below a certain threshold are pruned, and the model is fine-tuned to recover the drop in accuracy. This type of pruning is called unstructured pruning as the pruned connections have no specific pattern. This approach is useful when most of the parameters lie in the FC (fully connected) layers. Often, specialized libraries and hardware are required to leverage the induced sparsity to save computation and memory requirements. However, this does not typically result in any significant reduction in CNN computations (FLOPS based SpeedUp) as most of the calculations are performed in CONV (convolutional) layers. For example, in VGG-16, 90\% of the total parameters belong to FC layers, but they contribute to 1\% of the overall computations, which implies that CONV layers (having  10\% of the total model parameters) are responsible for 99\% of the overall calculations. 

Other works include \cite{chen2015compressing} where they propose hashing technique to randomly group the connection weights into a single bucket and then fine-tune the model to recover from the accuracy loss. 
\subsection{Filter Pruning}
In our work, we focus on filter level pruning. Most of the works in this category evaluate the importance of an entire filter and prune them based on some criteria followed by re-training to recover the accuracy drop. In the work  \cite{abbasi2017structural}, they calculate the filter importance by measuring the change in accuracy after pruning the filter from the model. \cite{li2016pruning} used $l_1$ norm to calculate the filter importance. \cite{hu2016network} calculate the filter importance on a subset of the training data using activation of the output feature map. These approaches are largely based on hand-crafted heuristics. Parallel to these works, ranking filters based on data-driven approaches are proposed. \cite{liu2017learning} performed the channel level pruning by attaching a learnable scaling factor to each channel and enforcing $l_1$ norm on those parameters during the training. Recently, group sparsity is also being explored for filter level pruning. \cite{alvarez2016learning,lebedev2016fast,wen2016learning,zhou2016less} explored the filter pruning using group lasso. However, at times these methods require specialized hardware for efficient SpeedUp during inference. 

Closest to our work is the work of \cite{molchanov2016pruning} where they proposed filter rankings using a mean of absolute gradient values and demonstrated that it gives competitive results to the brute-force method of checking loss deviation for each filter.

\subsection{Quantization}

Quantization based approaches aim to convert and store the network weights into a comparatively low bit configuration. The reduction in memory and computational requirements seems improbable after a certain level. However, these approaches can be used as a complement to filter pruning based approaches to extend the compression rates. Notably, \cite{han2015deep} compressed the model by combining pruning, quantization and Huffman coding. In the early works \, binarization \cite{binarycompression} has been used for the model compression. Extending this, \cite{zhou2016less} used ternary quantization learned from the given data. Recently, \cite{miao2017towards} conducted the network compression based on the float value quantization for model storage.  

At times, these quantization methods require specialized library/hardware support to reach desired compression rates. Some of the other notable works using different approaches from quantization include \cite{denton2014exploiting,zhang2015efficient} and \cite{jaderberg2014speeding} where they used the low-rank approximation to decompose tensors and reduce the computations.
 
Our method performs filter pruning using data-driven filter rankings. To the best of our knowledge, our work is a primary effort to relate filter importance to its stability and does not require any special hardware/software such as cuSPARSE (NVIDIA CUDA Sparse Matrix library). 

\section{Proposed Approach}
\begin{figure*}[t]
    \centering
    \includegraphics[scale=.5]{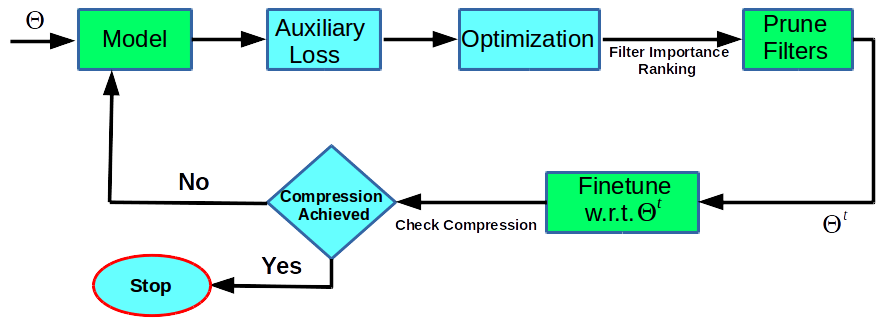}
    \caption{Stability based filter pruning approach where filters are pruned iteratively using auxiliary loss based optimization.}
    \label{fig:main}
\end{figure*}

\subsection{Terminology}
Let $\mathcal{L}_i$ denote the $i^{th}$ layer where $i\in [1,2,\dots K]$ and $K$ is the number of convolutional layers. The number of filters in layer $\mathcal{L}_i$  is represented by $n_i$ (which is also the number of output channels). The set of filters in layer $\mathcal{L}_i$ is denoted as $\mathcal{F}_{\mathcal{L}_i}$. Where $\mathcal{F}_{\mathcal{L}_i}=\{f_1^i,f_2^i,\dots, f_{n_i}^i\}$. The dimension of each filter $f_j$ is $(h_i,w_i,c_{in})$, where $h_i$, $w_i$ and $c_{in}(=n_{i-1})$ are height, width and number of input channels respectively.  $|f_j^i|$ denotes the sum of absolute values of filter $f_j^i$ and $f_{j,l}^i$ denotes the individual value of the filter.
\subsection{Approach}
In our work, we propose a filter pruning approach based on the stability of a filter, a metric which we use to measure the importance of a filter. The stability of a filter is inversely proportional to its sensitivity for a perturbation of loss function. The high sensitivity of a filter implies low stability (and hence less importance of that filter in the current task) and vice versa.
Let $C(\Theta)$  be the actual cost function of the model with model parameters $\Theta$. To create a perturbation, we introduce an auxiliary loss. This auxiliary loss is designed such that it forces the negative filter values to $-1$ and positive filter values to $+1$. The auxiliary loss for a layer i is given as:
\begin{align}\label{eq:plusminus}
S_{\mathcal{L}_i}=\sum_{j=1}^{n_i} \sum_{l=1}^{h_i*w_i*c_{in}} [(-1-f_{j,l}^i). \mathbb{I}(f_{j,l}^i< 0)\nonumber \\+ (1-f_{j,l}^i). \mathbb{I}(f_{j,l}^i\geq 0)]
\end{align}

where $\mathbb{I}()$ denotes the function which equals 1 if the condition is satisfied else 0. Now the complete loss can be given as:
\begin{equation}\label{eq:mainloss}
    L(\Theta)= C(\Theta)+ \lambda \sum_{i=1}^K S_{\mathcal{L}_i}
\end{equation}

Having defined the perturbation, we now describe the procedure for pruning.

\subsubsection{Training the network}
As we know that the deep networks have enough complexity to represent any function, it is quite possible that the auxiliary loss interferes with the optimization of an actual loss function. To avoid this possibility, we first train the network using actual loss function $C(\Theta)$. Let the filters at the end of training be denoted by $\mathcal{F}_{\mathcal{L}_i}=\{f_1^i,f_2^i,\dots, f_{n_i}^i\}$. We then train the network using the total loss function $L(\Theta)$. To avoid the possibility of drifting away from optimal weights for the actual task, we train only for the limited number of epochs (since the auxiliary loss is independent of data points, hence we use 1-3 epochs for optimizing equation-\ref{eq:mainloss}).
Let the $\mathcal{M}_{\mathcal{L}_i}=\{m_1^i,m_2^i,\dots, m_{n_i}^i\}$ be the set of filters at layer $\mathcal{L}_i$ after optimizing equation-\ref{eq:mainloss}.
\subsubsection{Ranking the filter Importance}
Once we have the $ \mathcal{F}_{\mathcal{L}_i}$ and $\mathcal{M}_{\mathcal{L}_i}$ for $C(\Theta)$ and $L(\Theta)$, we can calculate the importance of the filters in each layer $\mathcal{L}_i$. The filter ranking $(FI_{\mathcal{L}_i})$  of $\mathcal{L}_i$ is defined as the ratio of the sum of the absolute value of the filters after and before applying the auxiliary loss. This is given as:

\begin{equation}\label{eq:filterimportance}
    FI_{\mathcal{L}_i}= \left\{\frac{|m_j^i|}{|f_j^i|}:\forall j \in\{1,2,...,n_i \}\right\}
\end{equation}
The filter with the high ratio has high sensitivity and implies that it is an unimportant filter. The filter that has a strong contribution to the model has the least sensitivity, hence low ratio. Let $P=[p_1,p_2,\dots,p_K]$ be the number of filters to be pruned form each layer, where $K$ is the number of convolutional layers. Now based on the filter importance given by equation-\ref{eq:filterimportance}, select $p_1,p_2, \dots, p_K$ lowest important filters from respective layers in the model and prune them. The pruned set can be given as:
\begin{equation}
    P_{set}^t=\left\{\sigma_{p_1} ({\mathcal F_{L_1}}),\sigma_{p_2} ({\mathcal F_{L_2}}),\dots,\sigma_{p_k}({\mathcal F_{L_K}})\right\}
\end{equation}

Here $\sigma$  is the select operator that selects $p_i$ least important filters from the layer $\mathcal{L}_i$. $P_{set}^t$ is the set of filters that are discarded from the model.

\subsubsection{Pruning and fine-tuning}
Now we have the residual filters:
\begin{equation}\label{eq:leftparamter}
    \mathcal{F}^t=\mathcal{F}^{t-1} \setminus P_{set}^t
\end{equation}
$\mathcal{F}^t$ is the set of remaining filters in the model with parameters $\Theta^t$ after $t^{th}$ pruning iteration. $\setminus$ is the set-difference symbol. In each pruning iteration, after discarding the filter, we observe a small drop in accuracy. To avoid the accumulation of such accuracy drops, we fine-tune the residual network for 2-5 epochs. During fine-tuning, we use the actual loss (without auxiliary loss). We continue this procedure until the desired compression rate is achieved as shown in Figure-\ref{fig:main}.

\subsection{FLOPS and Memory Size Requirements}
In this section, we derive a formula for calculating FLOPS and Run Time Memory. 

For a convolutional neural network, the total number of computations (FLOPS) on the $\mathcal{L}_i$ convolutional layer can be given as:
\begin{equation}\label{flopconv}
    FLOPS_{conv_i}= c_{in}w_k h_k w_o h_o c_{o}*B
\end{equation}
Similarly, the total number of computations (FLOPS) on the $\mathcal{L}_i$ fully connected layer can be given as:
\begin{equation}\label{flopfc}
    FLOPS_{fc_i}= c_{in} c_{o}*B
\end{equation}
Where, $B$ is the batch size. ($w_{in},h_{in},c_{in}$), ($w_k,h_k,c_{in}$) and  ($w_o,h_o,c_o$) are the shapes of the input feature map (width, height, and number of input channels),  the convolution filter (width of kernel, height of kernel, and number of input channels) and  the output feature map (width, height, and number of output channels) respectively. 

Total FLOPS for the complete model can be given as:
\begin{equation}
   FLOPS=\sum_{i=1}^{K}  FLOPS_{{conv}_i} +\sum_{j=1}^{N}  FLOPS_{{fc}_j}
\end{equation}
Where $K, N$  is the number of convolutional and fully connected layers in the model respectively.

Run Time Memory (TRM) depends on the memory space created to store feature maps and model parameters. Hence total memory requirement for layer $\mathcal{L}_i$ can be estimated as:
\begin{equation}\label{mfeaturemap}
    M_{fm_i}=4w_{o}h_{o}c_{o}*B
\end{equation}
\begin{equation}\label{mweight}
    M_{w_i}= 4w_kh_kc_{in}c_o
\end{equation}
Where $M_{fm_i}$ is the memory space required to store the feature map ($w_o,h_o,c_o$) and $M_{w_i}$ is the memory space required to store parameters at layer $\mathcal{L}_i$.  Hence the total memory required for complete model (including all CONV and FC layers) can be given as:
\begin{equation}
    TRM=\sum_{i=1}^{K+N}  M_{fm_i} +\sum_{j=1}^{K+N}  M_{w_j}
\end{equation}

Note that, for FC layers, $w_k,h_k,w_o,h_o=1$ and $c_{in},c_o$ are the number of incoming and outgoing connections respectively. Similarly, for CONV layers $c_{in}, c_{o}$ are the number of input and output channels respectively. $M_{fm_i}$ linearly depends on the batch size ($B$). 

For the FLOPS computation, we ignore the cost of pooling, batch normalization, dropouts, etc., and the fused multiply-adds assumption is used. Inference time memory, which depends only on the feature maps and the weights, is reported.

\subsection{Relationship with the previous approaches}
In the work by Molchanov et.al. \cite{molchanov2016pruning}, they proposed to prune the channels using $\vert \Delta C(h_i) \vert (= \vert C(D, h_i=0)- C(D, h_i)\vert $) criteria. They used taylor series expansion to calculate the metric. They demonstrate the difference between their work and the Optimal Brain Damage (OBD) \cite{lecun1990optimal} by arguing that expected value of absolute value of a gaussian random variable is proportional to its variance and it serves as an important metric for pruning.

Their argument, in brief, states that after the completion of training, as per OBD,
\begin{equation}
\frac{\partial E_{x\sim p(x)}[C]}{\partial W_i} = E_{x\sim p(x)}\left [\frac{\partial C(x)}{\partial W_i}  \right ] = 0
\label{eq:1}
\end{equation}
Although, the expected value of the gradient of loss w.r.t a parameter, say $W_i$, may tend to zero, the individual samples need not have their cost function (cost function for that particular sample) indifferent to $W_i$. This is effectively captured in variance of $\frac{\partial C(x)}{\partial W_i}$. So, if the variance is higher then it is possible that the weight $W_i$ is indeed useful even though $E_{x\sim p(x)}\left [\frac{\partial C(x)}{\partial W_i}  \right ] = 0$. On the other hand, if the variance is low, then as the expectation also tends to 0, it is evident that the weight is useless and thus can be removed. Now, instead of calculating the variance explicitly, it suffices if we calculate 
\begin{align}
E_{x\sim p(x)}\left [ \left | \frac{\partial C(x)}{\partial W_i} \right | \right ]
\label{eq:2}
\end{align}As stated in \cite{molchanov2016pruning} this term is proportional to the variance of gradients over data distribution and hence can be used to rank filters. So, here they are making unconscious assumption that  $E_{x\sim p(x)}\left [\frac{\partial C(x)}{\partial W_i}  \right ] = 0$ when they start pruning

Let us call this assumption $A_1$ for the rest of the paper.
We first describe one scenario where the above method has issues with robustness. In their analysis, they considered that assumption $A_1$ holds. However, in practical scenarios, this may not hold true as practitioners follow different strategies such as early stopping, etc., where they stop training based on validation error. This implies that there is no guarantee that assumption $A_1$ holds for the training dataset. So, if we prune the weights according to equation-\ref{eq:2} in such scenarios, it may remove important weights since $E_{x\sim p(x)}\left [\frac{\partial C(x)}{\partial W_i}  \right ]$ may not be zero but $E_{x\sim p(x)}\left [ \left | \frac{\partial C(x)}{\partial W_i} \right | \right ]$ may be minimum. 

Hence, we propose a slight modification which attempts to remove the above disadvantage. We argue that as the networks are often over parametrized and it is obvious from the previous works that only a few of them contribute to an actual loss, the rest of the weights gets modified when an auxiliary loss function is added to existing loss function during the training. i.e.,  important weights for the actual task remain the same whereas the unimportant weights try to fit the auxiliary loss function when trained using both loss functions. We now formulate it mathematically.
\subsubsection*{Notation:}
Let the random variable X denote data distribution, and parameter W denote network weights and $\lambda$ be a scalar random variable. Let C denote the actual loss function, D denotes the auxiliary loss function, and L be the total loss function.
\subsubsection*{Formulation:}
The total loss function L is given by
\begin{equation}
L = C+ \lambda D
\label{eq:3}
\end{equation}
Now, the gradient of cost function w.r.t. a parameter, say $w_i$, depends on two random variables, X and $\lambda$. Since we argue that the important weights for actual task do not change due to the introduction of an auxiliary loss function, this implies that for a given data sample $X_i$, the following holds:
\begin{equation}
E_{\lambda}\left [\left |\frac{\partial L(X_i)}{\partial w_i} \right |\right] \approx   0
\end{equation}
To understand it better, compare it with the argument given by Molchanov et al., where the variance (over the data distribution) of the gradients w.r.t. unimportant weights will be low because they do not contribute to the loss function for the majority of the samples. Whereas the variance of gradients w.r.t. important weights will be high due to their contribution in loss function for all the samples. Here, we follow the same logic but with a minute change of taking the expectation over the joint probability distribution of ($X$,$\lambda$). Since the importance weights for the actual task are indifferent to auxiliary loss function (by our hypothesis), they contribute less to the update term during training with an auxiliary loss. So, when the $\lambda$ is varied, the resulting variance (of $\frac{\partial L}{\partial w_i}$) should be low. On the other hand, the unimportant weights for the actual task are the ones who try to fit the auxiliary loss function (by our hypothesis). So, when $\lambda$ is varied, the variance of $\frac{\partial L}{\partial w_i}$ will be high because when $\lambda=0$, $\frac{\partial L}{\partial w_i}$ =0 and when $\lambda\neq0$, $\frac{\partial L}{\partial w_i} \neq 0$ (by our hypothesis of unimportant weights). Hence resulting in a high mean and variance w.r.t. $\lambda$. As stated earlier, we do not train the network until the auxiliary loss is minimized as this may affect the actual task. This results in the above equalities being not perfectly satisfied and requires careful tweaking. But, as the gradients are proportional to change in weight values, we use the change in weight values criteria for pruning instead of mean of absolute gradient values. In practice, we found this approach to be effective.

\section{Results}
To evaluate our proposed work, we perform experiments on four standard models, LeNet-5 \cite{lecun1998gradient}, VGG-16 \cite{vgg2014very} and ResNet-50 \cite{resnet} for classification task and Faster-RCNN \cite{ren2015fasterrcnn} for object detection task. All the experiments are performed on TITAN GTX-1080 Ti GPU and i7-4770 CPU@3.40GHz. 

\begin{table}[!ht]
    \centering
    \scalebox{0.9}{
    \begin{tabular}{|c| c| c| c| c|} 
    \hline
    Method & Filter & Error\% & FLOPS & Pruned \% \\ [0.8ex] 
    \hline\hline
    Baseline & 20,50 & 0.83 & $4.40\times10^6$ & -- \\ 
    NIPS'16 \cite{wen2016learning} & 5,19 & 0.80 & $5.97\times 10^5$ & 86.42\\
    NIPS'16 \cite{wen2016learning} & 3,12 & 1.00 & $2.89\times 10^5$ & 93.42\\
    NIPS'17 \cite{neklyudov2017structured} & -- & 0.86 & -- & 90.47\\
    \hline
    \textbf{Prun-1 (ours)} &\textbf{4,14} &  \textbf{0.79}& $\mathbf{3.97 \times 10^5}$ & \textbf{90.98}\\
     \textbf{Prun-2 (ours)} &\textbf{3,8} &  \textbf{0.92}& $\mathbf{2.14 \times 10^5}$ & \textbf{95.14}\\
    \hline
    \end{tabular}
    }
    \caption{Table showing results for the LeNet-5 model on the MNIST dataset. SSL and SBP are proposed by \cite{wen2016learning} and \cite{neklyudov2017structured} respectively.}
    \label{tab:lenet5}
    
\end{table}

\subsection{LeNet-5 on MNIST}
MNIST is a data-set having 60,000 training images and 10,000 testing images. Two convolutional (20,50) and two fully connected layers (800,500) are present in the LeNet-5 model. We trained the model, and the trained model has an error rate of 0.83\%.

We optimized equation-\ref{eq:mainloss} for one epoch with $\lambda=0.00001$  to calculate filter importance in each pruning iteration. Learning rate is varied in the range $[0.001,0.0001]$ for this experiment. As compared to the previous approaches (Table-\ref{tab:lenet5}), we have a significantly higher FLOPS compression with the less drop in the accuracy. This proves the effectiveness of our proposed metric for filter ranking over the previous methods.

\begin{figure}[!ht]
    \centering
    \includegraphics[height=4.7cm, width=8.2cm]{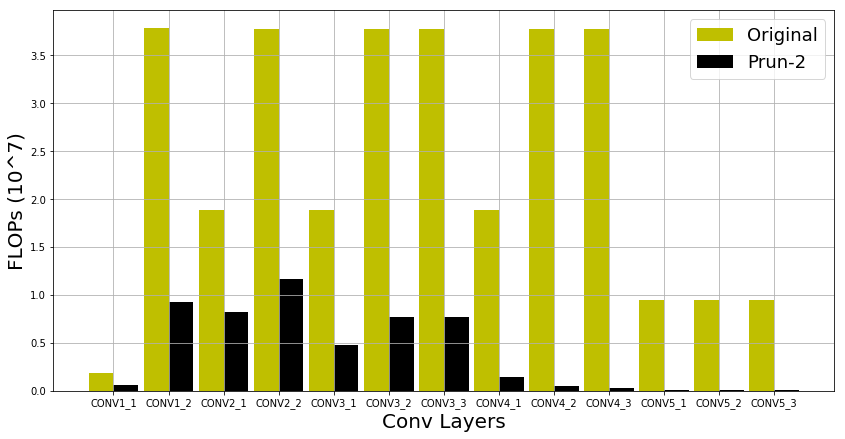}
   
    \caption{Figure shows the original and pruned model layer-wise FLOPS for the VGG-16 model on the CIFAR-10 data-set.}
    \label{fig:channel-FLOPS}
   
\end{figure}

\subsection{VGG-16 on CIFAR-10}
We perform an experiment on the VGG-16 model on the CIFAR-10 data-set. We use the same VGG-16 model and settings as mentioned in \cite{li2016pruning}, and after each layer, batch normalization is deployed. The model is trained from scratch. Layerwise FLOPS distribution is shown in Figure-\ref{fig:channel-FLOPS}. It is clear from Figure-\ref{fig:channel-FLOPS} that CONV1\_2, CONV2\_2, CONV3\_2, CONV3\_3, CONV4\_2, CONV4\_3 layers have much higher FLOPS as compared to the remaining layers. Hence, to compress FLOPS, we need to remove more filters from such layers. We optimized equation-\ref{eq:mainloss} for one/two epochs with $\lambda=0.00001$  to calculate filter importance ranking in each pruning iteration. We vary learning rate in the range $[0.001,0.0001]$ for this experiment. We get our first pruned model (Prun-1) after 82 epochs.

Table-\ref{tabvgglike} shows the detailed results for VGG-16 pruning. Table-\ref{tab:vgg-16} shows the comparison of our pruned model with previous approaches. Our method prunes 95.9\% of parameters on CIFAR10, significantly larger than 64.0\% pruned by \cite{li2016pruning}. Furthermore, our method reduces the FLOPS by 83.43\% compared to 34.2\% pruned by \cite{li2016pruning}. Layer-wise FLOPS distribution for  original and pruned model are shown in the Figure-\ref{fig:channel-FLOPS}. 
\begin{table}[ht]
    \centering
    \scalebox{0.77}{\begin{tabular}{|c|c|c|c|c|}
            \hline
            \multicolumn{2}{|c|}{} & \textbf{{Baseline}} & \textbf{{VGG-16 Prun1}} & \textbf{{VGG-16 Prun2}} \\ \hline
            \multicolumn{2}{|c|}{\textbf{Input Size}} & \textbf{32x32x3} & \textbf{32x32x3} & \textbf{32x32x3}  \\ \hline
            \multirow{16}{*}{Layers} & { CONV1\_1} &  64 &  31 &  20  \\ \cline{2-5} 
            & { CONV1\_2} & 64 &  53 &  50  \\ \cline{2-5} 
            & { CONV2\_1} & 128 & 84 &  71  \\ \cline{2-5} 
            & { CONV2\_2} & 128 & 84 &  71 \\ \cline{2-5}
            & { CONV3\_1} & 256 & 146 &  116 \\ \cline{2-5} 
            & { CONV3\_2} & 256 & 146 &  116 \\ \cline{2-5} 
            & { CONV3\_3} & 256 & 146 &  116 \\ \cline{2-5} 
            & { CONV4\_1} & 512 & 117 &  87  \\ \cline{2-5} 
            & { CONV4\_2} & 512 & 62 &  42 \\ \cline{2-5} 
            & { CONV4\_3} & 512 & 62 & 42 \\ \cline{2-5} 
            & { CONV5\_1} & 512 & 62 & 42 \\ \cline{2-5}
            & { CONV5\_2} & 512 & 62 & 42 \\ \cline{2-5}
            & { CONV5\_3} & 512 & 62 &  42 \\ \cline{2-5}
            & {FC6} & 512 & 512 & 512  \\ \cline{2-5} 
            & {FC7} & 10 & 10 & 10 \\ \hline
            \multicolumn{2}{|c|}{\textbf{Total parameters}} & 15.0M & 1.0M (15X)  & 0.62M (24.2X)  \\ \hline
            \multicolumn{2}{|c|}{\textbf{Model Size}} & 60.0 MB & 4.1 MB (14.6X) & 2.5 MB (24X)\\ \hline
            \multicolumn{2}{|c|}{\textbf{Accuracy}} & 93.49 & 93.43 & 93.02 \\ \hline
            \multicolumn{2}{|c|}{\textbf{FLOPS}} & 313.7M & 78.0M(4.02X) & 52.0M (6.03X) \\ \hline
        \end{tabular}}
        \caption{Table shows the layer-wise pruning results and pruned model details for VGG-16 model on CIFAR-10.}
        \label{tabvgglike}
        
    \end{table}
    
\begin{table}[!ht]
    \centering
    \scalebox{.82}{
    \begin{tabular}{|c| c| c| c|} 
    \hline
    Method &  Error\% & Param Pruned(\%) & FLOPS Pruned(\%) \\ [0.8ex] 
    \hline\hline
    \textbf{ICLR'17 \cite{li2016pruning}} &  6.60 & 64.0 &34.20\\
    \textbf{NIPS'17 \cite{neklyudov2017structured}} &  7.50 & -- & 56.52\\
    \textbf{NIPS'17 \cite{neklyudov2017structured}} & 9.00 & -- & 68.35\\
    \hline
    \textbf{Baseline} & 6.51& -- &-- \\
    \textbf{Prun-1 (ours)} & \textbf{6.57}&  \textbf{93.3} & \textbf{75.14} \\
    \textbf{Prun-2 (ours)} &  \textbf{6.98}&  \textbf{95.9} & \textbf{83.43} \\
    \hline
    \end{tabular}
    }
   \caption{Table shows the FLOPs pruning result for VGG-16 on the CIFAR-10 dataset. Weight-Sum and SBP are proposed by \cite{li2016pruning} and \cite{neklyudov2017structured} respectively.}
    \label{tab:vgg-16}
\end{table}

\begin{figure}[!h]
    \centering
    \includegraphics[height=6cm, width=8.4cm]{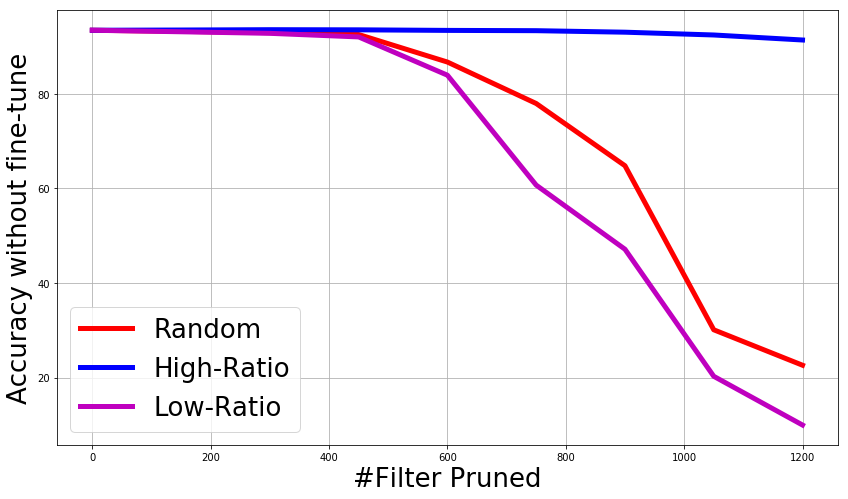}
    
    \caption{Effect of filter pruning with respect to accuracy for VGG-16. Filters are pruned from 6 Layers CONV4\_1 to CONV5\_3 simultaneously.}
    \label{fig:channel-pruning effect}
    
\end{figure}

\subsubsection{Ablation study on VGG-16}
We next show an ablation study on VGG-16 to demonstrate the effectiveness of the proposed filter importance ranking. Here, we pruned filters from 6 layers; Conv4\_1 to Conv5\_3 simultaneously. Since each layer from Conv4\_1 to Conv5\_3 contains 512 filters, therefore, a total of 512*6 filters are available for pruning. If we remove X filters in each layer from Conv4\_1 to Conv5\_3, then a total of 6*X filters gets pruned from the model.  Figure-\ref{fig:channel-pruning effect} horizontal axis shows the 6*X prune filters, and the vertical axis shows the accuracy without fine-tuning. We optimize equation-\ref{eq:mainloss} for three epochs with $\lambda=0.00001$  to calculate filter importance ranking. Figure-\ref{fig:channel-pruning effect} shows that if we prune filters from the low ratio (important filters), there is a sharp accuracy drop. A similar pattern is observed if we prune a filter randomly. In contrast, if we prune the filters from the high ratio (unimportant filters), then it results in a small accuracy drop even when we prune 1200 filters. 

\begin{figure}[t]
    \centering
    \includegraphics[height=5cm, width=8.4cm]{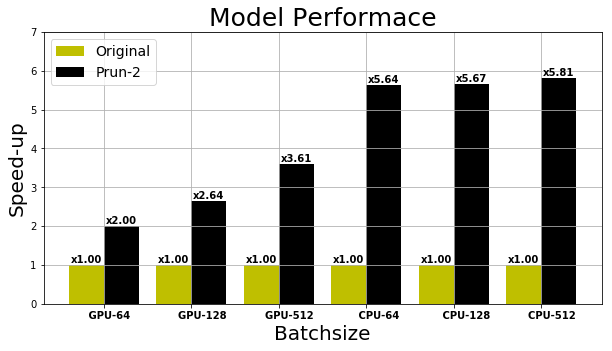}
    
    \caption{Figure shows the practical speed-up for the VGG-16 model on the CIFAR-10 data-set. Where i7-4770 CPU@3.40GHz CPU and TITAN GTX-1080 Ti GPU is used to calculate speed-up.}
    \label{fig:model_performance}
    
\end{figure}

\begin{table*}[!ht]
\centering
\scalebox{0.92}{
\begin{tabular}{|c|c|c|c|c|c|c|c|c|c|c|c|c|c|}
\hline
\multirow{2}{*}{\textbf{Model}}& \multirow{2}{*}{\textbf{data}} & \multicolumn{3}{c|}{\textbf{Avg. Precision, IoU:}}  & \multicolumn{3}{c|}{\textbf{Avg. Precision, Area }} & \multicolumn{3}{c|}{\textbf{Avg. Recall, \#Dets:}} & \multicolumn{3}{c|}{\textbf{Avg. Recall, Area:}}\\ \cline{3-14}
 & & \textbf{0.5:0.95} & \textbf{0.5} & \textbf{0.75}  & \textbf{S} & \textbf{M} & \textbf{L} & \textbf{1} & \textbf{10} & \textbf{100} & \textbf{S} & \textbf{M} & \textbf{L} \\ \hline
\textbf{F-RCNN original} & trainval35K & 30.3 & 51.3 & 31.8  & 13.8 & 34.6 & 42.6  & 27.3 & 41.3 & 42.4 & 22.4 & 47.9 & 58.5 \\ \hline
\textbf{F-RCNN pruned} & trainval35K & 30.6 & 51.0 & 32.2  & 14.7 & 34.7 & 42.5 & 27.7 & 42.0 & 43.2 & 23.8 & 48.1 & 58.9\\ \hline
\end{tabular}
}
\caption{Table shows the generalization results for Faster-RCNN on the MS-COCO data-set. In Faster-RCNN, we use our pruned ResNet-50 model (ResNet-50-Prun\_1) as a base model.}
\label{tab:coco}
\end{table*}

\begin{figure}[!h]
    \centering
    \includegraphics[scale=0.32]{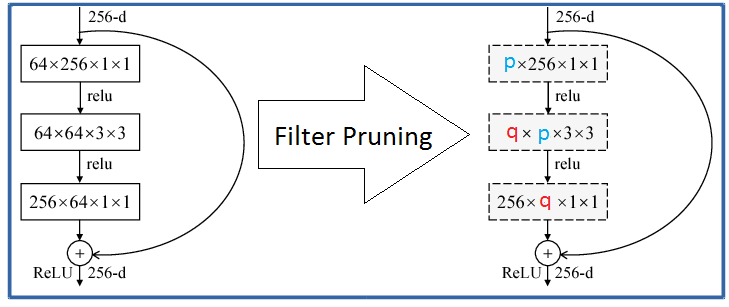}
    \caption{Figure shows our ResNet pruning strategy, where we perform pruning on the first two convolutional layers in each block to maintain the consistency over identity mapping.}
    \label{fig:resnet}
\end{figure}

\subsection{ResNet-50 on ImageNet}
We perform experiment on the large-scale ImageNet \cite{imagenet2015} data-set for the ResNet-50 model. The results are shown in the Table-\ref{resnettable} for the compressed model. Our pruned model achieved 44.45\% FLOPS compression while the previous method, ThiNet-70 \cite{luo2017thinet}, achieved  36.9\% FLOPS compression. Compared to ThiNet-70 we have significant better FLOPS compression.

Presence of identity mapping (skip connection) in ResNet model restrict pruning on the few layers. Since the output ($output=f(x)+x$) involves addition of $x$ and $f(x)$, hence $x$ and $f(x)$ need to be of same dimensions. This is the reason for pruning only two convolutional layers in each block as shown in Figure-\ref{fig:resnet}.

We pruned ResNet-50 from block 2a to 5c iteratively. The number of remaining filters from each layer in block 2, 3, 4 and 5 are 40, 80, 160 and 320 respectively in the pruned model. If a filter is pruned, then the corresponding channels in the batch-normalization layer and all dependencies to that filter are also removed.
We optimize equation-\ref{eq:mainloss} for one epoch with $\lambda=0.000005$  to calculate filter importance ranking in each pruning iteration. We vary learning rate in the range $[0.001,0.00001]$ for this experiment. Our pruned model (Prun-1) is obtained after 65 epochs.
Our results on ResNet pruning are shown in Table-\ref{resnettable}.

\begin{table}[!htb]
    \centering
    \scalebox{.77}{\begin{tabular}{|l|l|l|l|c|}
            \hline
            \textbf{Model} & \textbf{Top-5$\%$}  & \textbf{Parameters} & \textbf{FLOPS} & \textbf{Pruned FLOPS\%}\\ \hline
            \textbf{Baseline}  & 92.65 & 25.56M & 7.74G & --\\ 
            \textbf{ICCV'17 \cite{luo2017thinet}}       & 90.7 & 16.94M & 4.88G  & 36.9\\
            \textbf{IJCAI'18 \cite{he2018soft}}   & 92.0 &  --   &  --   & 41.8\\
            \textbf{Prun\_1 (ours)}  & \textbf{92.2} & \textbf{15.1M} & \textbf{4.3G} & \textbf{44.45}\\ \hline 
            
        \end{tabular}    }
    \caption{Table shows the comparison of our pruned model with \cite{luo2017thinet,he2018soft} for ResNet-50 FLOPS compression on the Imagenet data-set.}
    \label{resnettable}
    
\end{table}

\begin{figure}[t]
    \centering
    \includegraphics[height=5.5cm, width=8.4cm]{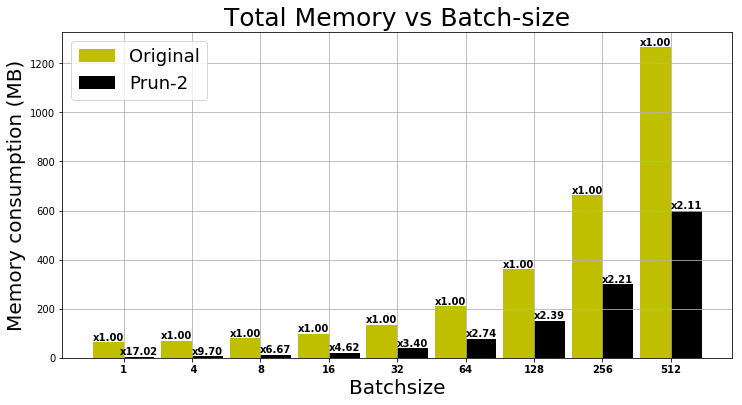}
   
    \caption{Figure shows Total Run Time (TRM) memory with respect to the batch size for VGG-16 models on CIFAR-10 data-set.}
    \label{fig:memory_size}
    
\end{figure}

\subsection{SpeedUp and Memory Size}
The theoretical FLOPS based SpeedUp is not the same as practical GPU/CPU SpeedUp. The practical SpeedUp depends on intermediate layers parallelization bottleneck, the speed of I/O data transfer, etc. TRM (Total Run-time Memory) depends on the number of parameters in the final compressed model, feature maps (FM) generated at run-time, batch-size (BS), the dynamic library used by Cuda, and all supporting header-file. But from the theoretical point of view, only model parameters size and feature maps size are considered in the TRM calculations. Hence TRM can be calculated as follows:
\begin{equation}
    TRM= MPS+ (FM*4*BS)
\end{equation}
Here we don`t have control over all the parameters barring model parameters size (MPS), FM and BS. We experiment VGG-16 on the CIFAR-10 dataset to show the practical SpeedUp and Memory size. SpeedUp and TRM results are shown in the Figure-\ref{fig:model_performance}, \ref{fig:memory_size} respectively.

As shown in the above equation, TRM grows linearly with respect to Batch size. Also, TRM  linearly depends on FM hence FM is the most critical factor for compressing the run-time memory. Filter pruning methods compress the model parameters as well as the depth of the feature maps hence filter level pruning methods achieves good compression for TRM. On the other hand, approaches based on inducing sparsity in the model only reduce the MPS and the size of the FM remains the same making batch size as the bottleneck. If we have constraints on batch size, this minimizes the parallelism on the GPU which results in a drop in speed. Figure-\ref{fig:memory_size}  explains that if we increase BS then TRM increases. Therefore we cannot afford large batches. The Figure-\ref{fig:model_performance} explains that for the small batch sizes, SpeedUp is degraded. Therefore for SpeedUp, we have to select a bigger BS, but then GPU or CPU memory bottleneck is there. Hence in the proposed method, we are pruning at filter level to compress FM memory.

The result for CPU and GPU SpeedUp over the different batch-size is shown in the Figure-\ref{fig:model_performance}. It is clear from the Figure-\ref{fig:model_performance} that with the increase in batch size, GPU has sharp SpeedUp, since on the small batch there it is not using its full parallelization capability. Although there are a lot of cores, only a few are used because the available data is limited whereas, on the bigger batch sizes, GPU uses its full parallelization capability. On the VGG-16 with 512 batch size, we have achieved 3.61X practical GPU SpeedUp while the FLOPS base theoretical SpeedUp is 6.03X. This gap is very close to CPU, and our approach gives the 5.81X practical CPU SpeedUp compare to 6.03X theoretical FLOPS base SpeedUp.

\subsection{Generalization Ability}
To show the generalization ability of our pruned model, we experimented on the object detection task. We are using the standard object detector Faster-RCNN \cite{ren2015fasterrcnn} on large-scale MS-COCO \cite{lin2014coco} data-set. We use ResNet-50 as the base network for Faster RCNN.

\subsubsection{Faster RCNN on COCO}

We performed experiments on the large-scale COCO detection
dataset which contain 80 object categories \cite{lin2014coco}. Here all the 80k
train images and a 35k val images are used for training (trainval35K) \cite{lin2017feature}. We are reporting the detection accuracies over the 5k unused validation images (also known as minival). We trained Faster-RCNN with the image-net pre-trained ResNet-50 as the base model to get F-RCNN original as shown in Table-\ref{tab:coco}.

For F-RCNN pruned, we used our pruned ResNet-50 model (Prun\_1) as given in Table-\ref{resnettable} as a base network in Faster-RCNN. It is clear from Table-\ref{tab:coco} that F-RCNN pruned model shows similar performance in all cases. However, some minor improvement in detection accuracies can be seen due to the reduction in over-fitting because of filter pruning. We used ROI Align and the stride  1 for the last block of the convolutional layer (layer4) in the base network (ResNet-50) in the Faster-RCNN implementation. Table-\ref{tab:coco} shows the results in detail.

\section{Conclusion}
In this work, we have proposed an approach to prune filters of convolutional neural networks and demonstrated a significant compression in terms of FLOPS and Run Time GPU memory footprint. To the best of our knowledge, our work is first of its kind in assessing filter importance using its robustness to auxiliary loss function. We have evaluated our method on various architectures like LeNet, VGG, and Resnet. Our method can be used in conjunction with other pruning methods such as binary/quantized weights to get further boost in SpeedUp. The experimental results show that our method achieves state-of-art results on LeNet, ResNet and VGG architecture. Moreover, we demonstrated that our pruning method generalizes well across tasks by pruning an architecture on one task and achieving competitive results using the same pruned model on another (but related) task. Additionally, the choice of auxiliary loss function plays an important role in compression for a certain task. This makes our approach flexible to adapt for a new task by selecting a data and task-dependent auxiliary loss function.
\newpage
{\small
\bibliographystyle{ieee}
\bibliography{egbib}
}

\end{document}